%% file: main.tex
\documentclass[]{jingdong}

\usepackage{float}
\usepackage{url}
\usepackage{amssymb}
\usepackage{tabularx}
\usepackage{array}
\usepackage{xspace}
\usepackage{listings}
\usepackage{enumitem}
\usepackage{etoc}
\usepackage{algpseudocode}
\usepackage{makecell}
\usepackage{booktabs}

\microtypesetup{expansion=false}
\renewenvironment{table*}{\begin{table}}{\end{table}}
\renewenvironment{figure*}{\begin{figure}}{\end{figure}}

\input{preamble}
\renewcommand{\pending}[1]{}

\title{Communicative On-Policy Distillation: Distilling Black-Box Teacher
Capabilities through Communication for Long-Horizon Tasks}

\author[1]{Lirui Luo}
\author[2]{Kelong Mao}
\author[3]{Xinwei Yang}
\author[4]{Rongqing Li}
\author[5]{Luyu Chen}
\author[2]{\protect\\[+0.18em]Kieran Wong}
\author[2]{Yudong Guo}
\author[2]{Xinrui Wang}
\author[6]{Jiayin Zhu}
\author[2]{Gu Simiu}
\author[2]{Sulong Xu}
\author[1]{Cong Fang}
\affiliation[1]{Peking University}
\affiliation[2]{JD.com}
\affiliation[3]{Sichuan University}
\affiliation[4]{Beijing Institute of Technology}
\affiliation[5]{Renmin University of China}
\affiliation[6]{The Hong Kong University of Science and Technology}

\checkdata[Project website]{\url{https://liruiluo.github.io/communicative-opsd/}}

\abstract{\input{sections/abstract}}

\begin{document}
\raggedbottom
\maketitle

\input{sections/intro}
\input{sections/related}
\input{sections/preliminaries}
\input{sections/method}

\input{sections/experimental_setup}
\input{sections/results}
\input{sections/conclusion}

\clearpage
\section*{LLM Usage Statement}
\noindent Large language model assistants (OpenAI GPT-6.1 Sol in Codex and DeepSeek-V4.1-Flash-1017 in the
DeepSeek Harness) were used to polish the writing, write experiment code, and help polish the figures and
tables. All model outputs were reviewed by the authors.

\bibliographystyle{ieeenat_fullname}
\bibliography{references}

\end{document}

%% file: preamble.tex
\newcommand{\pending}[1]{\textcolor{red!70!black}{[PENDING: #1]}}

\newcommand{\lPPO}{\mathcal{L}_{\mathrm{PPO}}}
\newcommand{\lOPSD}{\mathcal{L}_{\mathrm{OPSD}}}
\newcommand{\taup}{\tau^{+}}

\usepackage{amsthm}

\theoremstyle{definition}

\theoremstyle{plain}

\usepackage{listings}
\usepackage{array}
\usepackage{float}
\usepackage{placeins}
\usepackage{algpseudocode}
\floatstyle{ruled}
\newfloat{algorithm}{tbp}{loa}
\floatname{algorithm}{Algorithm}
\definecolor{casetitlebar}{RGB}{31,86,140}
\definecolor{casepromptblue}{RGB}{31,86,140}
\definecolor{caseboxbg}{RGB}{247,248,250}
\definecolor{caseboxrule}{RGB}{200,205,212}

\lstdefinestyle{taskExcerpt}{
  basicstyle=\ttfamily\scriptsize,
  breaklines=true,
  breakatwhitespace=false,
  columns=fullflexible,
  keepspaces=true,
  showstringspaces=false,
  backgroundcolor=\color{caseboxbg},
  frame=single,
  rulecolor=\color{caseboxrule},
  framexleftmargin=3pt,
  xleftmargin=3pt,
  xrightmargin=3pt,
  aboveskip=0pt,
  belowskip=2pt,
  moredelim=[l][\color{casepromptblue}\bfseries]{[TASK INPUT]},
}

%% file: sections/abstract.tex
On-policy distillation (OPD) is a powerful paradigm for improving
language-model agents on long-horizon tasks, yet its conventional
formulation relies on teacher logits that text-only black-box models
do not expose. To address this limitation, we introduce
\emph{Communicative On-Policy Distillation} (COPD), which uses
natural-language communication as a medium for distilling a stronger
model's capabilities into a smaller student. During task execution, the student autonomously decides when to consult a
black-box teacher and formulates open-ended natural-language requests without
predefined question templates or intervention schedules.
Successful student--teacher interactions are retained as privileged references to
supervise fresh adviser-free student rollouts through on-policy self-distillation,
jointly with task-reward reinforcement learning.
On CAMG-II, COPD achieves 2.46 times the gain of the strongest black-box OPD
baseline.
The capabilities it learns through communication
generalize to Terminal-Bench 2.1 and SWE-Bench Pro V2, improving task
success without adviser access.

%% file: sections/intro.tex
\section{Introduction}
\label{sec:intro}

\begin{figure*}[htbp]
  \centering
  \includegraphics[width=\textwidth]{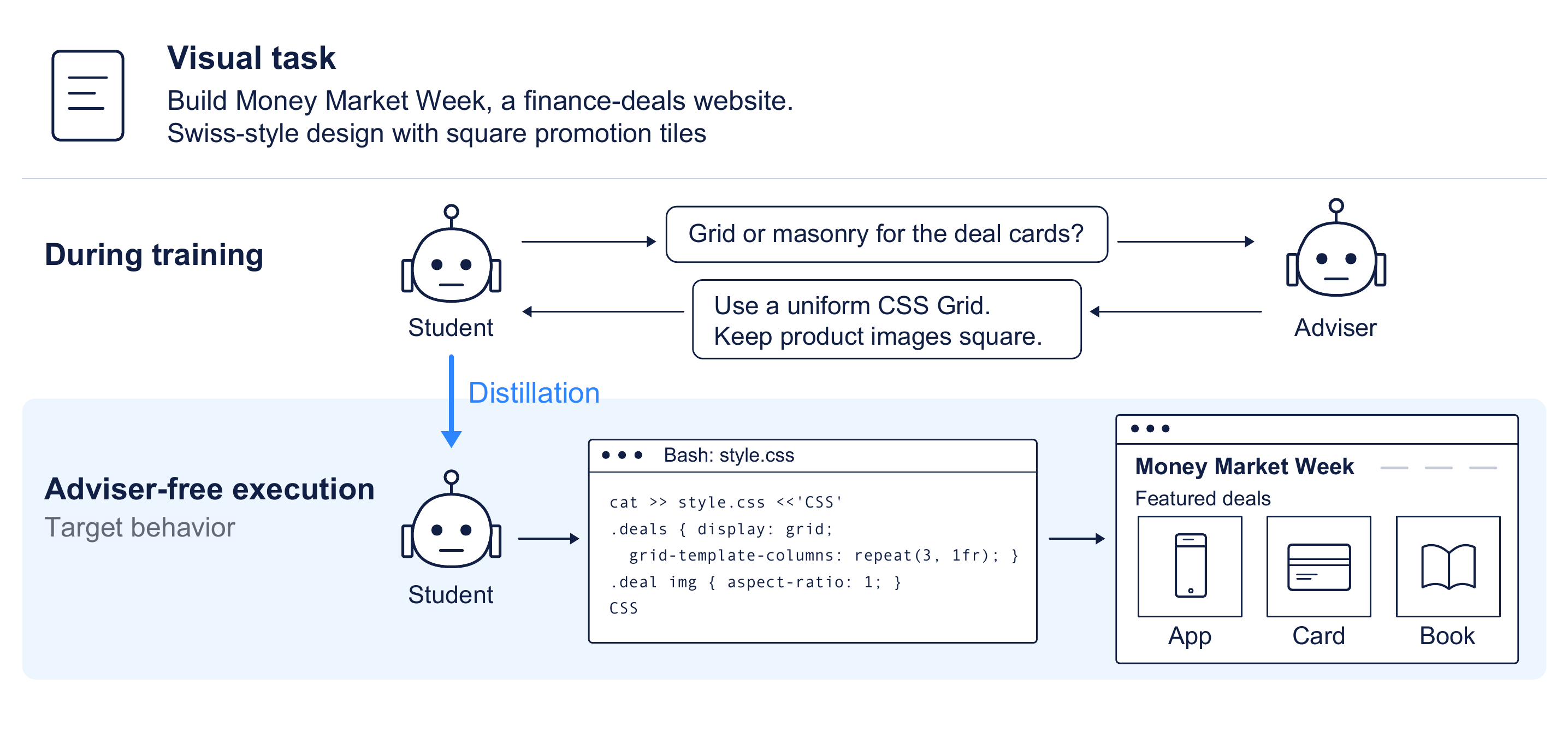}
  \caption{\textbf{Natural-language communication as a medium for distillation.} In a Visual task, the student asks a stronger black-box adviser whether to use a grid or masonry layout for a website's product cards. The adviser recommends a uniform CSS Grid with square product images. Through COPD, the student learns from successful consulted experience to act independently. The lower row illustrates the target adviser-free behavior.}
  \label{fig:teaser}
\end{figure*}

On-policy distillation (OPD) is a powerful paradigm for improving
language-model agents on long-horizon tasks
\citep{agarwal2024opd,omniopd2026}. Yet conventional OPD relies on
teacher logits, which text-only black-box models do not expose.
Human learning never needed logits: we consult those more experienced
and internalize their guidance to solve problems independently. Figure~\ref{fig:teaser} shows a concrete exchange:
a student building a website asks a stronger adviser how to organize
its page layout. Can a student agent learn to autonomously communicate with a
black-box teacher during task execution, and internalize the acquired knowledge
so that the teacher is no longer needed?

Acquiring useful guidance is itself nontrivial: in long-horizon tasks, neither
when the student will benefit from external expertise nor what information it
should communicate can be fully specified in advance. The student must therefore
learn to identify when to seek guidance and formulate task-specific messages as
execution unfolds, rather than relying on predefined intervention points or
question templates.

However, completing a task with advice does not ensure independent
execution once the adviser is removed. To bridge this gap, we introduce
\textbf{Communicative On-Policy Distillation (COPD)}, which uses
natural-language communication as a medium for distilling a stronger
model's capabilities into a smaller student. During task execution, the student autonomously decides when to consult a
black-box teacher and formulates open-ended natural-language requests without
predefined question templates or intervention schedules.
Successful student--teacher interactions are retained as privileged references to
supervise fresh adviser-free student rollouts through on-policy self-distillation,
jointly with task-reward reinforcement learning.

We train COPD-4B and COPD-9B from Qwen3.5 models of matching size on
CAMG-II~\citep{luo2026campii}. We evaluate independent execution without
adviser access.
On CAMG-II, COPD achieves 2.46 times the gain of the strongest black-box OPD
baseline.
The capabilities it learns through communication
generalize to Terminal-Bench 2.1 and SWE-Bench Pro V2, with improved
success rates.

Our contributions are as follows:
\begin{enumerate}
  \item \textbf{An algorithm for black-box distillation through natural
  language.} We introduce COPD, which trains a student to communicate
  with a stronger black-box adviser and internalize successful consulted
  experience through on-policy self-distillation for independent task
  execution (\S\ref{sec:method}).
  % Author-selected anticipated-results wording (2026-11-14).
  % Reconcile these two contribution claims with the completed experiments.
  \item \textbf{An analysis of what communication contributes.}
  We show that self-distillation internalizes knowledge conveyed through
  adviser replies, improving independent execution beyond
  communication-only training (\S\ref{sec:results}).
  \item \textbf{An extensive empirical evaluation across five domains.}
  We demonstrate that COPD improves independent task execution across
  diverse long-horizon tasks and generalizes beyond the training
  environments (\S\ref{sec:setup}, \S\ref{sec:results}).
\end{enumerate}

%% file: sections/related.tex
\section{Related Work}
\label{sec:related}

\paragraph{Learning to seek assistance.}
These methods train an agent to use an adviser that is present while
it acts. It learns what to ask through predicate questions
\citep{askbeforeact2022} or free-form clarification \citep{asktoact2025},
and whether to request assistance at all \citep{hilbench2026}. Guidance
can also follow failure, with discounted rewards for assisted successes
\citep{socraticpo2026}. These methods study help-seeking within specific assistance settings,
including oracle questions, task clarification, and failure-triggered guidance.
In contrast, COPD allows the student to initiate open-ended natural-language
communication at self-selected points during task execution and distills the
resulting successful interactions into an adviser-free policy.

\paragraph{Black-box on-policy distillation.}
Black-box on-policy distillation keeps OPD's student-generated
prefixes \citep{agarwal2024opd} and queries the teacher only for text.
Supervision must therefore come from its generations: chunk-level
verification against sampled teacher
continuations \citep{omniopd2026}, action-level rewards for semantic
agreement with teacher decisions \citep{yuan2026actionopd}, rubric
scores induced from teacher--student contrasts \citep{fang2026ropd},
an adversarial discriminator converted into an on-policy reward
\citep{ye2025gad}, and the teacher's completion of a student-generated
prefix as the training target \citep{wang2026olive}. These methods derive supervision
from black-box teacher outputs without training the student to initiate
communication during task execution. COPD instead enables the student to
autonomously seek guidance through free-form dialogue and uses successful
consulted experience to supervise its own fresh adviser-free trajectories.

\paragraph{On-policy self-distillation.}
On-policy self-distillation (OPSD) supervises student-generated
prefixes with a self-teacher conditioned on training-only
information. The information may be reference solutions
\citep{zhao2026self}, retrospective execution feedback or successful
rollouts \citep{hubotter2026sdpo}, or complete student trajectories
used to adapt the self-teacher \citep{feng2026past}. These methods construct privileged supervision from reference solutions or
the student's own experience. COPD instead conditions its self-teacher on
task-verified successful trajectories collected through student-initiated,
free-form communication with a black-box adviser. This allows guidance
acquired through dialogue to supervise fresh adviser-free student responses
without requiring teacher logits.

%% file: sections/preliminaries.tex
\section{Preliminaries}
\label{sec:preliminaries}

\paragraph{Multi-turn agentic reinforcement learning.}
For task $x$ in environment $d$, policy $\pi_\theta$ samples a response
$a_t=(y_{t,1},\ldots,y_{t,L_t})$ of $L_t$ tokens from bounded visible
history $h_t$. Executing the response returns an observation and native
reward $r_t$. An episode $\tau$ contains $T$ responses and has return
$G(\tau)=\sum_{t=1}^{T}\gamma^{t-1}r_t$, where $\gamma\in[0,1]$ is the
discount factor. The environment also determines success $s(\tau)\in\{0,1\}$.
With task distribution $\mu$ and trajectory distribution $p_\theta$
induced by the policy, task-reward RL maximizes
\begin{equation}
 J(\theta)=\mathbb{E}_{(d,x)\sim\mu,\,\tau\sim p_\theta(\cdot\mid d,x)}
 [G(\tau)].
 \label{eq:task-objective}
\end{equation}

\paragraph{On-policy distillation.}
OPD provides token-level teacher supervision on student-generated responses
\citep{agarwal2024opd}. At context $c=(h_t,y_{t,<j})$, the student samples
$y\sim\pi_\theta(\cdot\mid c)$, and a teacher distribution $q(\cdot\mid c)$
scores the same token. For fixed $c$ and $q$, minimizing the reverse KL
$D_{\mathrm{KL}}(\pi_\theta\|q)$ gives a teacher--student log-ratio signal.
The corresponding sampled-token policy-gradient surrogate is
\begin{equation}
\begin{split}
A^{\mathrm{OPD}}_\theta(c,y)&=\operatorname{sg}\!\left[\log q(y\mid c)-\log\pi_\theta(y\mid c)\right],\\
\ell_{\mathrm{PG}}(\theta)&=-A^{\mathrm{OPD}}_\theta(c,y)\,\log\pi_\theta(y\mid c).
\end{split}
\label{eq:opd-general}
\end{equation}
During differentiation, the sampled $(c,y)$ and teacher distribution $q$
are held fixed, and $\operatorname{sg}$ stops gradients through the feedback.
On-policy self-distillation (OPSD) forms a teacher view of the student
conditioned on additional training-time information, such as a reference
solution or feedback \citep{zhao2026self,hubotter2026sdpo}.

%% file: sections/method.tex
\section{Method}
\label{sec:method}

\begin{figure*}[htbp]
  \centering
  \includegraphics[width=\textwidth]{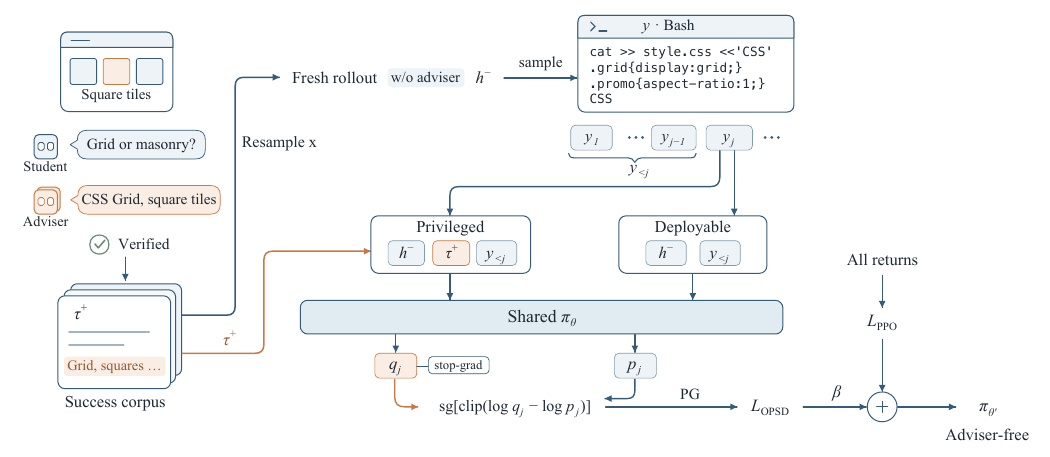}
  \caption{\textbf{Overview of COPD.} The student acquires guidance through
communication with a stronger black-box adviser. Successful consulted
trajectories enter a corpus, from which tasks are resampled for fresh
adviser-free rollouts. Two views of the current student score the same
sampled tokens, with only the privileged view receiving a successful
trajectory $\tau^+$ from the same task. Their log-probability difference
provides distillation feedback, which is combined with task-reward PPO to
train independent execution.}
  \label{fig:architecture}
\end{figure*}

In this section, we present COPD for distilling a stronger black-box
model's capabilities into a smaller student through language
(Figure~\ref{fig:architecture}). Inspired by human learning through
communication, we first collect successful consulted experience
(\S\ref{sec:method:protocol}), then use it to supervise fresh
adviser-free responses (\S\ref{sec:method:roles}), and finally combine
distillation with task reward to train independent execution~(\S\ref{sec:method:loss}).
These components are jointly optimized throughout training rather than implemented
as separate training stages.

\subsection{Collecting Experience through Communication}
\label{sec:method:protocol}

To access the stronger adviser's task-solving knowledge, we first let the
student communicate with it during task execution. The collection policy
$\pi_b$ is a snapshot of the student $\pi_\theta$ and generates episodes
with adviser access (w.~adviser) and without adviser access (w/o~adviser).
Since useful guidance depends on the task state, communication is itself a
policy action. At every step at which interaction is permitted, the student
either acts in the environment or sends free text to the adviser,
\begin{equation}
  a_t \in \mathcal{A}_{\mathrm{env}} \cup \mathcal{A}_{\mathrm{comm}},
  \qquad
  a_t^{\mathrm{comm}} = \operatorname{Ask}(u_t),
  \quad u_t \sim \pi_\theta(\cdot \mid h_t),
  \label{eq:comm-action}
\end{equation}
where $u_t$ is text the student generates from its own history, not a template
question, a predefined query type, or a choice among fixed options. The student
submits it through an ordinary \texttt{bash} command and receives the reply as a
text observation that guides subsequent actions. The adviser answers in text and
never acts in the environment, and the student may communicate again at any later
step. Native task rewards train the request tokens together with the policy
responses that follow them, so the communication policy and the task policy are
one model with one parameter set $\pi_\theta$, learned jointly from environment
returns rather than through a separately trained query selector.

To turn these interactions into distillation data, we retain an execution
only if it passes the native success check $s(\tau)=1$ and actually used
the adviser, denoted by $\operatorname{consulted}(\tau)$. Each reference
$\taup$ contains the complete trajectory of requests, replies, actions,
and observations, preserving how guidance was used during successful
execution. We store these references in a rolling corpus $\mathcal C$
indexed by environment and task $(d,x)$. They provide the context for
supervising fresh adviser-free attempts.

\input{sections/algorithm}

\subsection{Constructing Privileged Supervision}
\label{sec:method:roles}

To internalize the guidance for independent execution, we apply on-policy
self-distillation (OPSD) to fresh adviser-free responses. In standard OPSD the
privileged information may be a reference answer, execution feedback, or other
additional context. Here it is a complete, environment-verified trajectory that
the student itself produced while communicating during an earlier attempt. It
requires neither logits from the black-box adviser nor target actions for the
student, and the adviser is never asked to judge whether a particular token of a
fresh rollout is correct. It conversed with the student during a real task, and
the student's own privileged view turns that conversation into a token-level
learning signal. We therefore condition the current student on a successful
consulted execution to form a privileged self-teacher
\citep{zhao2026self,hubotter2026sdpo}. The recorded choices and
outcomes provide task information for assessing unaided responses.

To construct these training pairs, we mix scheduled tasks with tasks
resampled from $\mathcal C$ and execute each
selected task anew in a fresh workspace. Each resulting w/o~adviser attempt
can be paired with an available consulted success of the same task.
Corpus sampling thus creates pairing opportunities even when tasks occur only
once in the source schedule. 

Since a fresh attempt may reach different states from the reference, we
score its responses along its own unassisted history. For example, the
successful CSS Grid implementation in Figure~\ref{fig:architecture} provides
context for scoring a fresh response that sets the layout and aspect ratio.
For episode $i$, response $t$, and token $j$, let
$c_{i,t,j}=(h_{i,t}^{-},y_{i,t,<j})$ be the unassisted history and sampled
response prefix. Given the paired success $\tau_i^{+}$, the two views are
\begin{align}
 p_{i,t,j} &= \pi_\theta(\cdot\mid c_{i,t,j}), \\
 q_{i,t,j} &= \operatorname{sg}\!\left[
   \pi_\theta(\cdot\mid c_{i,t,j},\tau_i^{+})\right],
 \label{eq:copsd-views}
\end{align}
where $p$ is the deployable view and $q$ is the detached privileged view.
Both share current student weights and the same unassisted prefix, while
only $q$ receives $\tau_i^{+}$. The privileged view scores the student's sampled
response without generating another action \citep{agarwal2024opd}.
The external adviser participates only during collection.

\subsection{Joint Reinforcement Learning and Distillation}
\label{sec:method:loss}

Finally, we train the student with native task rewards and feedback from
the privileged view. For each sampled token $y=y_{i,t,j}$, the distillation
feedback is the clipped log-probability gap
\begin{equation}
 A^{\mathrm{OPD}}_{i,t,j}=\operatorname{sg}\!\left[
 \operatorname{clip}\!\left(\log q_{i,t,j}(y)-\log p_{i,t,j}(y),-C,C\right)
 \right],
 \label{eq:copsd-token-feedback}
\end{equation}
where $C>0$ bounds the feedback. A positive value encourages a token that
the privileged view finds more likely, while a negative value discourages
it.

Because asynchronous collection can use older student weights,
we retain the actual behavior probability $b_{i,t,j}(y)$ recorded at
generation and form $\rho=p_{i,t,j}(y)/b_{i,t,j}(y)$. This ratio reweights
sampled tokens, while the two current-weight views supply the feedback. The clipped surrogate corresponding to Eq.~\ref{eq:opd-general},
with token indices omitted, is
\begin{equation}
 \ell^{\mathrm{OPD}}=-\min\!\left(\rho A^{\mathrm{OPD}},\,
 \operatorname{clip}(\rho,1-\epsilon,1+\epsilon)A^{\mathrm{OPD}}\right).
 \label{eq:copsd-token-pg}
\end{equation}
Here $\epsilon$ is the ratio-clip width.

Aggregating valid paired w/o~adviser token losses into $\lOPSD$,
we jointly optimize
\begin{equation}
 \mathcal L(\theta)=\lPPO(\theta)+\beta\,\lOPSD(\theta),
 \label{eq:copsd-objective}
\end{equation}
where $\beta\ge0$ is the distillation weight. Task PPO~\citep{schulman2017ppo}
trains execution and communication from native returns on both w.~adviser
and w/o~adviser episodes, including failures. OPSD uses paired w/o~adviser
responses to internalize consulted experience for independent execution.

Algorithm~\ref{alg:copsd} summarizes the update, with asynchronous collection.

%% file: sections/algorithm.tex
\begin{algorithm}[t]
\caption{Communicative On-Policy Distillation}
\label{alg:copsd}
\small
\begin{algorithmic}[1]
\Statex \textbf{Input:} initial student weights $\theta_0$, training tasks $\mathcal T$,
\Statex \hspace{\algorithmicindent} and fixed black-box adviser $\mathcal A$
\State $\theta\gets\theta_0$, rolling reference corpus $\mathcal C\gets\emptyset$
\While{training continues}
  \State $\mathcal X\gets$ \Call{SampleTasks}{$\mathcal T,\mathcal C$} \Comment{task batch}
  \State $\mathcal D\gets$ \Call{Rollout}{$\pi_b,\mathcal X,\mathcal A$} \Comment{completed episodes}
  \State $\mathcal S\gets\{\tau\in\mathcal D:s(\tau)=1\land\operatorname{consulted}(\tau)\}$ \Comment{successes}
  \State $\mathcal C\gets$ \Call{RetainRecent}{$\mathcal C\cup\mathcal S$} \Comment{corpus window}
  \For{each w/o~adviser trajectory $\tau_i^-\in\mathcal D$}
    \State $\tau_i^+\gets$ \Call{Retrieve}{$\tau_i^-,\mathcal S,\mathcal C$} \Comment{same-task reference}
  \EndFor
  \For{optimizer minibatch $\mathcal B$ from $\mathcal D$}
    \State Compute native-task $\lPPO$ on all eligible rows
    \State $\lOPSD\gets0$
    \For{each valid paired w/o~adviser token $y$ in $\mathcal B$}
      \State Score $y$ with student $p$, privileged $q$ at $\theta$ (Eq.~\ref{eq:copsd-views})
      \State Compute feedback $A^{\mathrm{OPD}}$ (Eq.~\ref{eq:copsd-token-feedback})
      \State Compute token loss $\ell^{\mathrm{OPD}}$ (Eq.~\ref{eq:copsd-token-pg})
    \EndFor
    \State Aggregate $\lOPSD$ with the fixed token denominator
    \State Update $\theta$ with $\lPPO+\beta\lOPSD$ (Eq.~\ref{eq:copsd-objective})
    \State Update the critic from native returns
  \EndFor
\EndWhile
\Statex \textbf{Output:} adviser-free student $\pi_\theta$
\end{algorithmic}
\end{algorithm}

%% file: sections/experimental_setup.tex
\section{Experiments}
\label{sec:experiments}

We evaluate whether COPD improves the student's ability to execute tasks
without an adviser. We first compare task performance with task-reward
controls and black-box distillation methods. We then examine what
communication contributes, using component ablations, matched reference
sources, and reply-content interventions. Communication dynamics,
test-time adviser access, and the training communication budget connect
the acquisition of guidance to independent execution. External benchmarks
assess transfer beyond the training environments.

\subsection{Experimental Setup}
\label{sec:setup}

\paragraph{Task environments.}
\label{sec:setup:substrate}

We use Coding Agent Memory Gym II (CAMG-II)~\citep{luo2026campii},
which spans Shop, Coding, AutoResearch, General, and Visual tasks.

\paragraph{Training.}
We train COPD from fresh \texttt{Qwen3.5-4B} and
\texttt{Qwen3.5-9B} students~\citep{qwen2026qwen35} for 100 training updates
with 64 episodes per update, totaling 6,400 episodes, under an
equal mixture over the five domains. Training is fully asynchronous,
combining native task-reward PPO with OPSD and using a fixed
\texttt{DeepSeek-V4.1-Flash} adviser.

\paragraph{Baselines.}
As the task-reward reference, we use CAMP-I~\citep{luo2026camg},
which learns with the same file memory and no adviser access.
We also compare five black-box OPD methods: learned discriminator
rewards in GAD~\citep{ye2025gad}, rubric rewards in
ROPD~\citep{fang2026ropd}, supervision on teacher continuations of student
prefixes in OLIVE~\citep{wang2026olive}, rewards for semantic agreement
with teacher actions in ActionOPD~\citep{yuan2026actionopd}, and token weighting through
speculative verification in OmniOPD~\citep{omniopd2026}.
For GAD, we include variants without and with teacher-response warm-up.
The main comparison uses the same 4B initialization and evaluation
protocol.

\paragraph{Ablations.}
We compare COPD with w/o OPSD, w/o communication, and w/ SFT.
The w/o OPSD variant removes the distillation update while retaining
communication and task-reward PPO.
The w/o communication variant disables the adviser during training and uses
the student's verified unassisted successes as references for OPSD on fresh
attempts at the same tasks. The w/ SFT variant replaces OPSD with supervised
fine-tuning (SFT) on the student's responses in successful consulted
trajectories, retaining communication and task-reward PPO.

\paragraph{Evaluation.}
We report per-domain scores and their equal-weight average, with the native metric of each domain. All methods use the same
CAMG-II test set of 128 tasks per domain, giving 640 tasks in total,
with adviser access disabled in the main comparison.
We measure external transfer on the network-free task subsets of
Terminal-Bench 2.1 and SWE-Bench Pro
V2~\citep{merrill2026terminalbench,swebenchprov2}.

%% file: sections/results.tex
\begin{table*}[htbp]
\centering
\small
\begin{tabular*}{\textwidth}{@{\extracolsep{\fill}}lcccccc@{}}
\toprule
 & \multicolumn{5}{c}{Task success (\%)} & \\
\cmidrule(lr){2-6}
Method & Shop & Coding & AutoResearch & General & Visual & Avg.$\uparrow$ \\
\midrule
\multicolumn{7}{l}{\textit{Frozen control}} \\
Qwen3.5-4B~\citep{qwen2026qwen35}               & 77.3 & 27.3 & 46.1 & 39.8 & 36.0 & 45.3 \\
Qwen3.5-9B~\citep{qwen2026qwen35}               & 90.6 & 26.6 & 56.2 & 60.9 & 36.5 & 54.2 \\
\midrule
CAMP-I~\citep{luo2026camg} & 85.2 & 32.0 & 38.3 & \textbf{68.8} & 46.5 & 54.2 \\
GAD~\citep{ye2025gad}                      & 0.0 & 18.8 & 7.0 & 35.2 & 16.6 & 15.5 \\
GAD (w/ warm-up)~\citep{ye2025gad}         & 85.2 & 22.7 & 39.8 & 57.8 & 39.0 & 48.9 \\
ROPD~\citep{fang2026ropd}                     & 73.4 & 27.3 & 43.0 & 63.3 & 40.8 & 49.6 \\
OLIVE~\citep{wang2026olive}                     & 79.7 & 24.2 & 39.8 & 53.1 & 39.0 & 47.2 \\
ActionOPD~\citep{yuan2026actionopd} & \textbf{93.0} & 24.2 & 41.4 & 58.6 & 39.0 & 51.2 \\
OmniOPD~\citep{omniopd2026}                   & 80.5 & 25.8 & 43.0 & 55.5 & 40.8 & 49.1 \\
COPD (ours)              & 92.2 & \textbf{36.7} & \textbf{55.5} & 66.4 & \textbf{48.2} & \textbf{59.8} \\
COPD-9B & \textbf{93.0} & \textbf{38.3} & \textbf{57.0} & \textbf{68.8} & \textbf{49.4} & \textbf{61.3} \\
\midrule
\midrule
Teacher & 100.0 & 78.1 & 83.6 & 91.4 & 50.4 & 80.7 \\
\bottomrule
\end{tabular*}
\caption{\textbf{COPD attains the highest average success rate among the trained
policies.} Among the 4B systems it outperforms CAMP-I, every black-box OPD
baseline, and the frozen student. The largest gain lands in General, 26.6
points over the frozen student, and Shop and General are the two cells where a
baseline stays ahead. Scaling the same recipe to a 9B student lifts the average
to 61.3. Visual reports the mean judged grade.}
\label{tab:primary}
\end{table*}

\subsection{Results}
\label{sec:results}

Table~\ref{tab:primary} reports adviser-free task performance. The
following analyses take that gain apart: which components produce it, what
the student keeps when the adviser is gone, and how its communication
changes during training.

\paragraph{COPD improves independent execution across domains.}
\label{sec:results:primary}

To test whether communicating with a stronger adviser improves what the
student can do on its own, we evaluate every policy on the same CAMG-II
panels with adviser access disabled (Table~\ref{tab:primary}). COPD-4B
raises the mean adviser-free task score from 45.3 to 59.8, against 54.2 for
CAMP-I. The 14.5-point gain is 2.46 times what ActionOPD, the strongest of
the five black-box OPD baselines, obtains at 51.2. Applied unchanged to the
9B student, the recipe reaches 61.3, seven points above its 54.2 starting point.

Two properties of the update distinguish COPD from the other black-box
distillation baselines and may account for the lead: the trajectory is the
student's own, and the signal arrives at every token. The teacher contributes
language and never an action, so every supervised token is sampled from the
current policy and the update stays on-policy, whereas OLIVE and OmniOPD
supervise segments the teacher wrote and those segments lie off the student's
policy. Two views of the same tokens give a signal at every position, whereas
GAD and ROPD reduce an episode to one discriminator or rubric score and
ActionOPD pays out only at the turns it groups.

\paragraph{COPD internalizes the adviser's guidance.}
\label{sec:results:adviser-access}

\begin{figure}[htbp]
\centering
\includegraphics[width=0.5\textwidth]{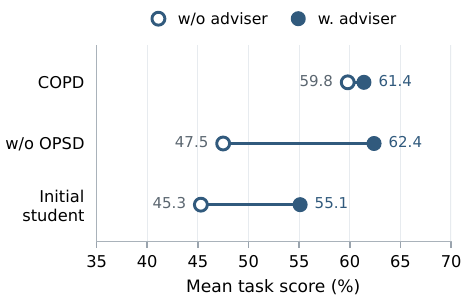}
\caption{\textbf{COPD internalizes most of the help it receives.} Each line spans
one policy's two conditions, adviser removed and adviser available, on the same
tasks. Access is worth 9.8 points to the untrained student and 14.9 to w/o OPSD,
and 1.6 to COPD-4B, which is the only policy whose level barely moves with the
adviser.}
\label{fig:adviser-access}
\end{figure}

To test whether what the student gains is internalized guidance rather than
a dependence, we score each policy on the same tasks twice, with the adviser
reachable and with it removed (Figure~\ref{fig:adviser-access}). Removing it costs the initial student a fifth of its score, w/o OPSD nearly a
quarter, and COPD-4B a fortieth. The ordering flips with access: w/o OPSD passes
COPD-4B at 62.4 against 61.4 while the adviser is there, and falls 12.3 points
behind it without.

\paragraph{A consulted success teaches more than an unassisted one.}
\label{sec:results:reference-utility}

\begin{figure}[htbp]
\centering
\includegraphics[width=0.5\textwidth]{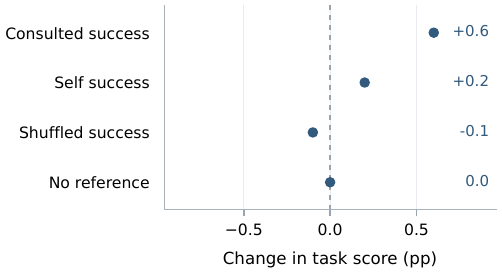}
\caption{\textbf{The same success is worth more when it comes from a consultation.} We
start from one student checkpoint and allow a single OPSD update, varying only
the reference: a consulted success of the task, the student's own unassisted
success, a consulted success from another task in the same domain, or nothing.
Each updated policy is then scored on a test set, and the horizontal
axis is the resulting change in adviser-free task score. The unassisted success
moves that score, the consulted success of the same task moves it several times
further, and a success taken from another task moves it slightly the other
way.}
\label{fig:reference-utility}
\end{figure}

The capability survives without the adviser, so what put it there is the material
the student distilled. To separate a reference's content from the number of
successes it supplies, we compare consulted and unassisted references on tasks
where both succeed, holding the student, the fresh attempt, and the update fixed.
The consulted success moves the score 0.6 points on a test set, three times the
0.2 of the student's own unassisted success (Figure~\ref{fig:reference-utility}).
A consulted success carried over from another task moves it the other way, by
$-0.1$, and no reference leaves the score at 0.0. The guidance therefore carries
the update, and it has to fit the task: a success from another task is worth
less than none.

\paragraph{Changing the replies changes what the student learns.}
\label{sec:results:reply-content}

\begin{figure}[htbp]
\centering
\includegraphics[width=0.5\textwidth]{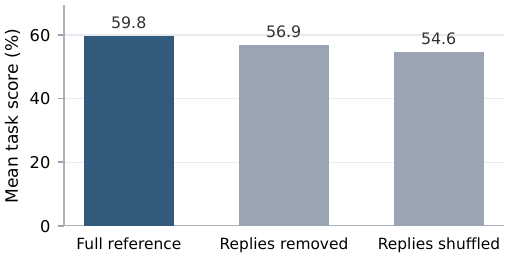}
\caption{\textbf{The adviser's words are worth more than the solution they came with.} We
run the full recipe three times, changing only the replies inside the privileged
reference: the complete success, the same success with the words deleted, and
those words moved onto a success from another task. Mean adviser-free success falls
from 59.8 to 56.9 when the words are deleted and to 54.6 when they come from
another task, so the words help only when they describe the task being learned.}
\label{fig:reply-content}
\end{figure}

To test how much of the supervision sits in the replies' content, we retrain the
recipe twice more: once with the replies deleted, and once with them moved onto a
success from another task, keeping the questions, actions, and observations
unchanged (Figure~\ref{fig:reply-content}). Removing the replies costs 2.9 points and misplacing them 5.2, which takes COPD-4B
to 54.6, level with the task-reward reference. The replies therefore add supervision of
their own, and they add it only when they fit the task.

\paragraph{The student asks less often and asks much better.}
\label{sec:results:consultation}

\begin{figure}[htbp]
\centering
\includegraphics[width=0.5\textwidth]{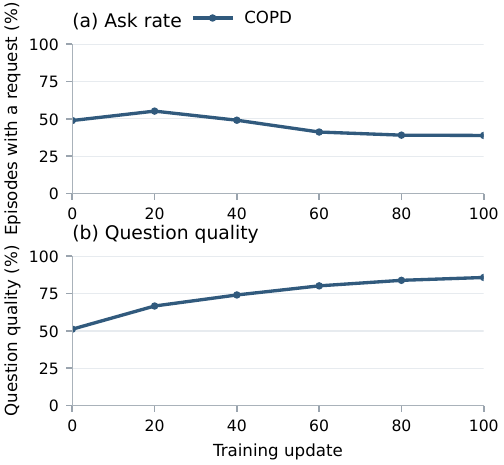}
\caption{\textbf{The student ends training asking less often and asking far
better.} Ask rate and question quality for COPD, measured every twenty updates.
Ask rate is the fraction of adviser-enabled episodes containing at least one
request. Question quality is a rubric score on task relevance, clarity, and
context sufficiency, given by gpt-6-astra from the request and the student's
pre-request context, normalized to 0--100. Requests peak at update 20 and then
fall, while question quality climbs throughout, so the student leans on the
channel less and uses it better.}
\label{fig:consultation}
\end{figure}

To test whether the student's use of the channel changes as the policy improves, we
measure its ask rate and the quality of its questions every twenty updates
(Figure~\ref{fig:consultation}). Communication carries the guidance that COPD
distills, so a student that asks less and asks better extracts more from each
exchange. Requests rise from 48.9\% at the start to 55.2\% at update 20, then fall
to 38.9\% by update 100. Quality moves the other way: gpt-6-astra's scores climb
from 51.2 to 85.8, most of the rise in the first forty updates. The student
therefore ends training asking less often and asking far better, spending the
channel where the task needs it instead of at every opportunity.

\paragraph{Ablation study.}
\label{sec:results:ablations}

\begin{figure}[htbp]
\centering
\includegraphics[width=0.5\textwidth]{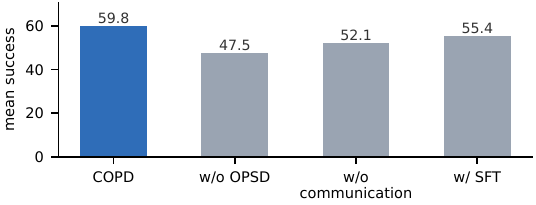}
\caption{\textbf{Every component carries part of the gain.} Mean adviser-free
success over the five domains for COPD and three variants that each remove one
piece: the distillation update, the adviser during training, and on-policy
targets replaced by supervised fine-tuning on the consulted responses. The w/o
communication variant keeps distillation and references the student's own
verified unassisted successes. Every variant falls short, and the largest drop
comes from the update itself.}
\label{fig:ablations}
\end{figure}

To test whether each piece of the recipe is necessary, we remove one piece at a
time and score every arm without adviser access on the same test set
(Figure~\ref{fig:ablations}). The w/o OPSD variant keeps communication and
task-reward PPO but drops the distillation update, and reaches 47.5, 12.3
points below the full recipe and below CAMP-I at 54.2, so consulted experience
does not become capability on its own. The w/ SFT variant imitates the consulted
responses instead of scoring the student's own attempts, and reaches 55.4, 4.4
points short, so the target of the update matters as much as its presence. The
w/o communication variant keeps distillation but references the student's own
verified unassisted successes, and reaches 52.1, 7.7 points short, so the
reference has to come from consulted experience. The update carries the largest
share of the gain, and neither the channel alone nor imitation of the same
trajectories substitutes for it.
\paragraph{The acquired capability transfers beyond CAMG-II.}
\label{sec:results:external}

We evaluate the initial students, a larger same-generation model, and both trained
policies on Terminal-Bench 2.1 and SWE-Bench Pro V2 without an adviser
(Table~\ref{tab:external}). COPD-4B gains 4.9 points on Terminal-Bench and 5.3 on
SWE-Bench Pro, and COPD-9B gains 9.9 and 4.4. Both policies move on
both suites, so the gains are not concentrated in one. At 30.3 the 9B policy comes
within 0.3 points of Qwen3.5-35B-A3B at 30.6, a larger model of the same
generation.

\begin{table}[htbp]
\centering
\scriptsize
\setlength{\tabcolsep}{1.8pt}
\renewcommand{\arraystretch}{1.06}
\caption{\textbf{Both trained policies improve on both suites, and COPD-9B
closes to within 0.3 points of a larger same-generation model.} Resolution
rate on Terminal-Bench 2.1 and SWE-Bench Pro V2 for the initial students, a
larger same-generation model, and the trained policies.}
\label{tab:external}
\begin{tabular}{@{}lccc@{}}
\toprule
Policy & TB 2.1 & SWE-Bench Pro V2 & Avg. \\
\midrule
Qwen3.5-4B & 23.1 & 11.2 & 17.1 \\
Qwen3.5-9B & 23.1 & 23.1 & 23.1 \\
Qwen3.5-35B-A3B & \textbf{33.8} & 27.4 & \textbf{30.6} \\
\midrule
COPD-4B & 28.0 & 16.5 & 22.3 \\
COPD-9B & 33.0 & \textbf{27.5} & 30.3 \\
\bottomrule
\end{tabular}
\end{table}

\paragraph{Case study.}
\label{sec:results:case-study}

\begin{figure*}[htbp]
\centering
\includegraphics[width=\textwidth]{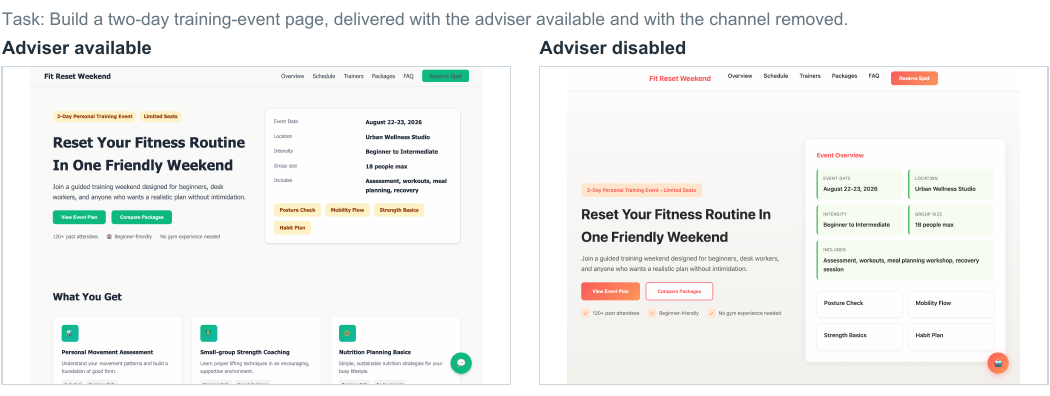}
\caption{\textbf{The trained policy reaches a comparable page with or without an
adviser.} The same visual brief is delivered twice, once with an adviser mounted
and advertised and once with the channel removed. The two pages differ in detail
but are comparable in structure and finish, so the behavior the consultation
established does not depend on reaching an adviser at test time.}
\label{fig:case-study}
\end{figure*}

Figure~\ref{fig:case-study} takes one Visual brief, a two-day training-event page,
and runs it twice: once with the adviser mounted and advertised, once with the
channel removed. Both runs return a finished page with the same sections in a
comparable layout, so the behavior established during consultation is available
without the channel. This is the qualitative counterpart of the aggregate
adviser-free Visual score in Table~\ref{tab:primary}.

\FloatBarrier

%% file: sections/conclusion.tex
\section{Conclusion}
\label{sec:conclusion}

In this paper, we introduced Communicative On-Policy Distillation (COPD), a
framework for transferring black-box teacher capabilities through
student-initiated, open-ended natural-language communication. Rather than
prescribing when or how the student should seek guidance, COPD allows the student
to communicate autonomously during task execution and internalizes successful
consulted experience through on-policy self-distillation. Across five long-horizon
domains, COPD substantially improves adviser-free task performance over existing
black-box distillation methods, with gains transferring to Terminal-Bench 2.1 and
SWE-Bench Pro V2. Our ablations and controlled interventions show that successful
communication alone does not guarantee independent capability: task-relevant
guidance must be internalized into the student's own policy. These findings
establish autonomous communication as a viable mechanism for black-box knowledge
transfer, enabling students not only to learn with stronger teachers, but also to
learn from those interactions to act independently.

\paragraph{Limitations.}
The study is limited to students up to Qwen3.5-9B and to distillation from a single frozen
teacher. Whether the observed gains persist at larger scales or carry over to multi-teacher OPD
(MOPD) remains untested.